\documentclass[format=sigconf, screen=true, review=false]{style/acmart}
\usepackage{graphicx}

\renewcommand\footnotetextcopyrightpermission[1]{} 
\AtBeginDocument{%
  \providecommand\BibTeX{{%
    \normalfont B\kern-0.5em{\scshape i\kern-0.25em b}\kern-0.8em\TeX}}}

\begin{document}

\title{Annotation Techniques for Judo Combat Phase Classification from Tournament Footage}

\newcommand{\AuthorInfo}[2]{
    \author{#1}
    \affiliation{
        \institution{Georgia Institute of Technology}
        \city{Atlanta}
        \country{USA}
    }
    \email{#2}
}

\AuthorInfo{Anthony Miyaguchi}{acmiyaguchi@gatech.edu}
\AuthorInfo{Jed Moutahir}{jmoutahir3@gatech.edu}
\AuthorInfo{Tanmay Sutar}{tanmay21@gatech.edu}



\maketitle

\section{Abstract}

This paper presents a semi-supervised approach to extracting and analyzing combat phases in judo tournaments using live-streamed footage.
The objective is to automate the annotation and summarization of live streamed judo matches.
We train models that extract relevant entities and classify combat phases from fixed-perspective judo recordings.
We employ semi-supervised methods to address limited labeled data in the domain.
We build a model of combat phases via transfer learning from a fine-tuned object detector to classify the presence, activity, and standing state of the match.
We evaluate our approach on a dataset of 19 thirty-second judo clips, achieving an F1 score on a $20\%$ test hold-out of 0.66, 0.78, and 0.87 for the three classes, respectively.
Our results show initial promise for automating more complex information retrieval tasks using rigorous methods with limited labeled data. 
\section{Introduction}

Judo is a martial art and combat sport that focuses on throwing and grappling.
It was created by Jigoro Kano in 1882 and debuted at the 1964 Olympics.
Judo is principled on mutual benefit and maximum force with minimum effort through grappling competition.
Judo tournaments allow athletes to test their technique with resistant opponents.
Tournaments sort competitors into weight divisions and follow rules that are typically consistent with the Olympic rule set.
Large judo tournaments are now often live-streamed to a global audience.

In this paper, we explore annotation techniques to analyze judo competition footage for combat phase classification through semi-supervised identification of players and referees.
As a combat sport, Judo competitions follow a regular pattern that can be divided into smaller components and phases.
Competitors are pooled together in a bracket, and each match is a one-on-one contest.
The matches can then be divided into combat phases such as bowing, standing, and groundwork.
Entire tournaments are captured on video, but tasks such as match segmentation and highlight extraction are often done manually.
Annotation techniques can be used to automate these tasks and provide valuable insights into the competition.

\begin{figure}[t]
    \centering
    \includegraphics[width=\linewidth]{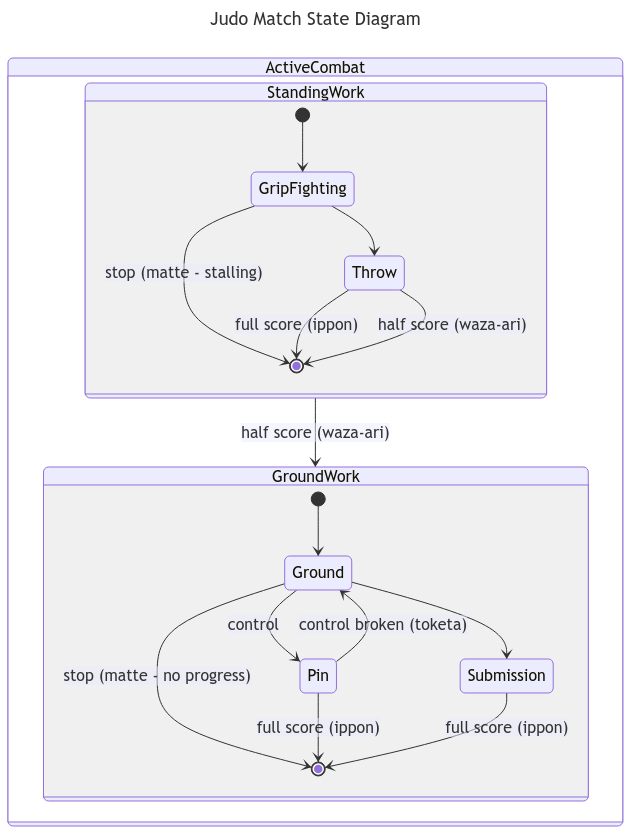}
    \caption{
        A state diagram of the active portions of Judo combat.
        The timer is actively running and is delineated by calls from the referee.
        Most active combat occurs during the standing portion of the match (tachiwaza), where players attempt to unbalance and throw each other.
        The match may continue to the ground (newaza) if a throw is not decisive.
        The state diagram is missing matches that end due to disqualification, such as executing a banned or dangerous technique.
    }
    \label{fig:active-combat-phases}
\end{figure}

\section{Related Work}



\subsection{Object and Pose Detection}

Object detection and pose detection has been studied for decades.
Detectron2 \cite{detectron2} provides a generic framework for object inference tasks, including region classification and pose detection.
\citet{quinn2022automation} combine real-time object detection, tracking, and pose detection to generate performance statistics from boxing and MMA footage.
They experiment with contact counters for punches and kicks by calculating intersection over union statistics using YOLOv5 for tracking object regions.
\citet{hudovernik2022video} build an automated jiu-jitsu scoring pipeline with a tracking classifier over keypoint poses and visual embedding features.
They classify combat positions by static frame, using a moving average to add temporal coherence.
\citet{ludwig2023all} propose a methodology to generate human pose estimation using a limited set of key-points generated from 27 hours of triple, high, and long jump competitions.
They are able to obtain state-of-the art performance with 2403 images annotated with 20 key-points using DensePose to generate segmentation masks.

\subsection{Human Activity Recognition}

Activity recognition is a widely studied topic within the multimedia literature, and there exists literature focusing on combat art.
We would like to be able to identify scored techniques by their Kodokan Judo name.
These techniques often happen in a dynamic sequence of actions to setup a technique that are executed in the span of a second. 

Video activity recognition models are often tested against large databases of catalogued videos.
The Sports-1M \cite{karpathy2014large} and Kinetics \cite{carreira_quo_2017} are large activity recognition datasets collected from YouTube.
Many authors will train their models on these large datasets, and test one-shot performance against the smaller HMDB \cite{kuehne2011hmdb} and UCF-101 \cite{soomro2012ucf101} datasets.

Deep learning has proven effective in human activity recognition, building on the success of convolutional neural networks and transformers in the vision domain through temporal extensions.
\citet{tran_closer_2018} study 3D CNNs for spatio-temporal action recognition by augmenting deep residual networked with \text{(2+1)D} layers representing the decomposed 3D convolution reaching state of the art performance.
Vision transformers (ViT) have become popular due to their ability for self-supervised learning and ability to scale with data.
\citet{liu2022video} develop a Video Swin Transformer by adapting the model to accept spatiotemporal tokens in the form of a cube.
\cite{tong2022videomae} extend the learning efficiency of video transformers through an aggressive (90-95\%) masking ratio of embedding tokens due to temporal redundancy of videos.

\subsection{Judo Time-Motion Analysis}

Judo has been studied under the lens of time-motion analysis using human exports to annotate video footage.
LINCE \cite{gabin2012lince} is one-such software for conducting analysis of sports footage.
\citet{marcon2010structural} statistically summarize combat into six phases (break, preparation, grip, technique, fall, and groundwork) using annotations done by domain experts on video footage.
They collected 276 action sequences during active match time with an average of 11 actions per match. 
\citet{stankovic_course_2015} analyze 140 matches from the men's division of the 2011 Paris World Championship. 
They categorize videos into actions (effective and ineffective attacks, penalties, and groundwork), and further categorize actions into gripping stance and mode of attack (leg/foot, hip, hand, and sacrifice).
They summarize their results by weight class into actionable coaching advice.  
\citet{soriano2019time} perform another time motion analysis of 150 judoka in 181 matches focused on the gripping portion of combat, revealing differences between sex and weight class.
They categorized 59.3\% of active match time to "trying" or "established" gripping time.
\citet{franchini2013judo} survey the literature regarding time-motion analysis and physiology, summarizing key effort and pause periods across various studies.
They uncover an effort-pause ratio between 2:1 and 3:1 during matches with 20s and 30s effort periods.
Automating analysis from footage could help with characterizing the direction of the sport with respect to rule changes, as well as informing players of effective strategies during matches. 

\section{Judo Combat State Model}

We describe the state of a judo match as transitions in the relative physical positioning of the players and the referee.
Each match starts and ends with the players bowing to their partner.
The match will start and swap between an active combat phase with a ticking timer and an intermission where players will reset into a neutral position.
The active combat sections consist of two primary states: a standing state, where players are fighting for control and execution of a throwing technique, and a ground state, where players are fighting for a pin or submission.
When enough points or penalties have accumulated, the refereed will make a decision, and the players will bow out of the match.
This state diagram in Figure \ref{fig:active-combat-phases} describes active combat, but a full tournament might be realized by a larger number of states.

\section{Dataset}

We obtained approximately 110 hours of live judo footage from an official USA Judo tournament called the President's Cup in November 2023.
The footage consists of ten live-stream views of the playing area for the duration of the tournament in a fixed perspective, capturing mats that two players and referees occupy.
The footage does not include markers for the time locations of the beginning and end of matches.
Most of the footage includes a software overlay that provides information about the players, such as their weight bracket, match time, and current score.
Each file is approximately 5.5GB.

For annotation purposes, each video is treated as a sequence of images.
An hour of footage from each mat is selected at one frame per second, providing a balance between the number of frames and the distribution of action for labeling.
For this particular dataset, footage offset by an hour is preferred due to a large number of empty or unusual frames from the tournament opening ceremony.

\section{Methodology}

Combat phase information from the tournament footage is extracted in several stages.
First, each frame is classified by whether a match is currently taking place.
This enables filtering out irrelevant content and providing anchoring points for video segmentation.
Next, the players and referees bounding boxes are detected with a deep learning model.
Finally, a combat phase model is constructed using transfer learning of the trained entity detector with lagged features over a sliding window of frames.

YOLOv8 \cite{Jocher_Ultralytics_YOLO_2023} is used as the primary object classification and detection model for this study.
Label Studio \cite{Label} is used for human annotation workflows.

\subsection{Full-Scene Match Classification}

The goal of full-scene match classification is to filter out irrelevant frames and to identify individual matches begin and end.
A labeling workflow is established using Label Studio images randomly selected from the training dataset.
Annotators are briefed on the semantic meanings of each label.
The frames are sorted into four classes: "match", "no match", "match intro", and "match outro". 
An ongoing match is defined by the presence of two players and a referee, typically delineated at the beginning and end by players bowing to each other.
A frame is not in a match state when there is a transition between players and is typically observed as the referee standing alone on the mat.
The software overlay defines the match intro and outro states, where player names and scores are displayed at the beginning and end of the match.

We use these labels to segment the videos into individual matches by finding transitions between match and no-match states and match intro and outro states.
A YOLOv8 classifier is fine-tuned against ground truth labels using a 70/15/15 train/validation/test split with the F1 score as the primary metric.

\subsection{Player and Referee Entity Detection}

\begin{figure}
    \centering
    \includegraphics[width=\linewidth]{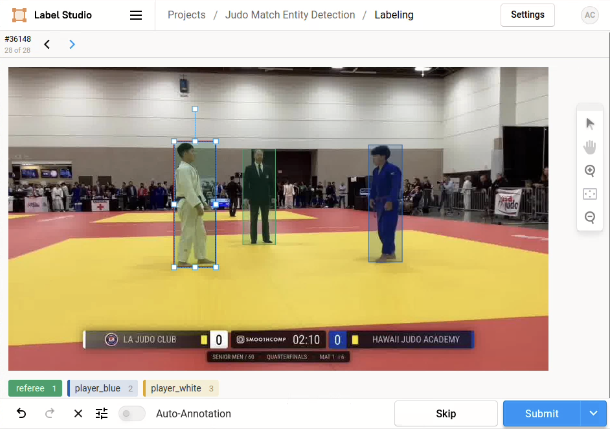}
    \caption{
        The bounding box of each player and referee is pre-annotated using rules derived from an object detector in a Label Studio project.
        Human annotators manually correct and validate the results.
    }
    \label{fig:label-studio-entities}
\end{figure}

Entity detection involves identifying the players and referees in the foreground of the video.
The first phase of entity recognition involves placing bounding boxes around the referee and the players to build a higher-order understanding of the scene.

Frames from the training dataset are manually labeled to form ground truth.
Annotators must draw a bounding box for the player in white, the player in blue, and the referee.
Players are designated by the color of their uniform: white and blue.
Players must wear a judo-gi in their respective color at higher-level tournaments.
Referees traditionally wear a black suit.
The entire bounding box includes occluded parts, which is often the case in close-quarter combat. 

YOLOv8 is fine-tuned to detect bounding boxes around all people and objects within the frame.
Pre-annotations are generated to assist with manual labeling using YOLOv8 to find all potential objects in a given scene as per figure \ref{fig:label-studio-entities}.
The top three objects are sorted by descending area and then assigned to a white player, a blue player, and a referee class based on the majority vote of pixels on each pixel from white, navy blue, and black, respectively.
The annotation dataset is divided into a 70/15/15 ratio for training, validation, and testing, respectively.
An object detector is fine-tuned to distinguish between multiple entities on a ground-truth test set.
The model can be used for downstream tasks, including filtering out irrelevant scenes containing the wrong number of participants or transfer learning via embeddings in the layers before the detection head.

\subsection{Combat Phase Classification}

The combat phase classifier determines the current phase of the match.
We use the player bounding boxes as a proxy determine the current phase of active combat.
In particular, we hypothesize that a particularly well-trained player and referee entity detection model learns a representation of images that is amenable classification into three classes with logical dependencies.

We manually annotate 30-second intervals of competition footage into three classes.
We label whether the mat is currently running a match, whether it is active, and if the players are standing.
Each label depends on the previous label, forming a decision tree.
We perform 19 annotations and then split these into a train and test dataset.
We generate labels for the classification task by quantizing into shorter 1-second intervals (Table \ref{table:phase_label_distribution}), where the goal is to predict each of the three binary classes at each second.

\begin{figure}
    \centering
    \includegraphics[width=\linewidth]{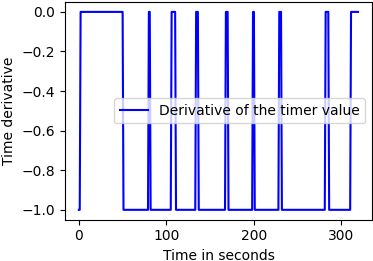}
    \caption{
       Match timing information from the overlay is extracted using Tesseract OCR \cite{smith2007overview}.
       Gaps from invalid readings are interpolated.
       The derivative of the timer is computed and plotted.
       When the derivative is 0, the timer is paused; when it is -1, it runs.
       In this match, the timer was paused nine times: at the beginning, seven pauses by the referee, and end of the match.
    }
    \label{fig:example-timer-extraction}
\end{figure}

As a heuristic for pre-annotation, we use a combination of extracted timer derivatives and the bounding-box aspect ratio to determine the presence of the match, the activity of the match and the standing state given by Figure \ref{fig:active-combat-phases}.
We get the derivative of the timer by extracting data from the tournament software overlay containing the overall match time and current score.
We extract the timer value using Tesseract \cite{smith2007overview} on the specific region of the screen where the timer is displayed.
We then use the derivative of the timer value to directly infer whether a match is present and active (see Figure \ref{fig:example-timer-extraction}).
We obtain bounding box information about players from the entity detection model and build two heuristics to determine the standing state of players.
If the player's bounding boxes are tall and skinny, they are more likely to stand.
If the player's bounding boxes are short and wide, they are more likely to be on the ground.

We fit supervised classification models to predict the three binary targets.
Our simplest model uses the second-to-last embedding layer of the pre-trained and fine-tuned YOLOv8 detectors as features in a logistic regression model trained separately for each target.
We also compare these results against the detector embeddings using the discrete cosine transform (DCT) to reduce the dimensions via low-pass filtering of the coefficients.
We choose 8, 16, 32, and 64 coefficients for the 1D DCT.
Given the different dimensions of the pre-trained and fine-tuned, we use the N-D DCT to reduce the dimensions for each embedding size into the same number of coefficients. 
The embeddings, 1D and N-D DCT coefficients are treated as individual feature representations. 
We also test modeling with features lagged by time $t \in \{1,2,3,5\}$.
Action in judo matches happens in seconds, so we do not need a large context window.
We concatenate the current frame's features with the features from the previous frames.
\section{Results}



\subsection{Full-Scene Match Classification}

We perform 2314 annotations across three annotators.
The dataset is imbalanced, with a large majority of frames being classified as "Match" and a small fraction of frames being classified as "Match Intro" or "Match Outro".
This matches the distribution of frames of the live-footage, as you would expect the majority of the footage to be dominated by the match itself and not the intermissions.

\begin{table}[h]
    \centering
    \caption{
        Distribution of full-scene match classes on a ground-truth human-annotated dataset of randomly sampled images.
        The subset of annotations from annotator \#1 are reported.
    }
    \begin{tabular}{|l|r|r|}
    \hline
    \textbf{category} & \textbf{images} & \textbf{percent} \\ \hline
    no match & 308 & 34.0\%  \\
    match & 446 & 49.2\% \\ 
    match intro & 128 & 14.1\% \\
    match outro & 25 & 2.7 \% \\ \hline
    total & 907 & \\ \hline
    \end{tabular}
    \label{tab:match_train_data_stats}
\end{table}

\begin{table}[h]
    \centering
    \caption{
        Validation results of fine-tuning YOLOv8 for full-scene scene match classification.
        Scores are reported with a 70-15-15 train-validation-test split.
    }
    \begin{tabular}{|l|l|l|l|l|}
    \hline
    \textbf{Category} & \textbf{F1-score} & \textbf{Accuracy} &\textbf{Precision} &\textbf{Recall} \\ \hline
    no match & 0.9714 & 0.9826 & 0.9623 & 0.9808 \\ \hline
    match & 0.9833 & 0.9827 & 0.9860 & 0.9806 \\ \hline
    match intro & 0.9877 & 0.9971 & 1.0000 & 0.9756 \\ \hline
    match outro & 1.0000 & 1.0000 & 1.0000 & 1.0000 \\ \hline
    \end{tabular}
    \label{tab:match_classification_train}
\end{table}

\begin{figure}[t]
    \centering
    \includegraphics[width=\columnwidth]{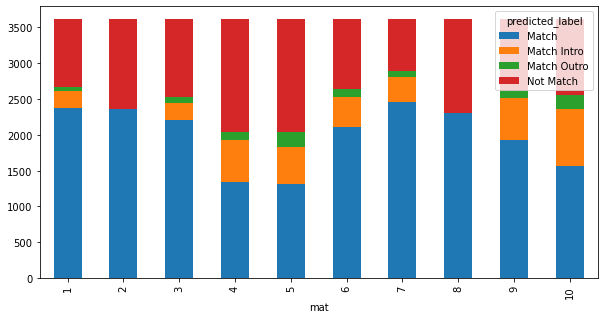}
    \caption{
        The distribution of full-scene match classes in the inference on the training set.
        Mats 2 and 8 do not have a video overlay.
    }
    \label{fig:match_classifier_histogram_infer}
\end{figure}

We train a YOLOv8 classifier and report the results in Table \ref{tab:match_classification_train}.
For evaluation, we run the model against the entire pool of frames used for annotation.
We see a large majority of frames being classified into the "Match" class, as per Figure \ref{fig:match_classifier_histogram_infer}.

\begin{figure}[t]
    \centering
    \includegraphics[width=\columnwidth]{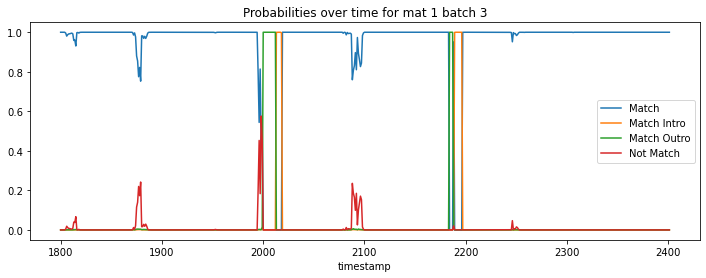}
    \includegraphics[width=\columnwidth]{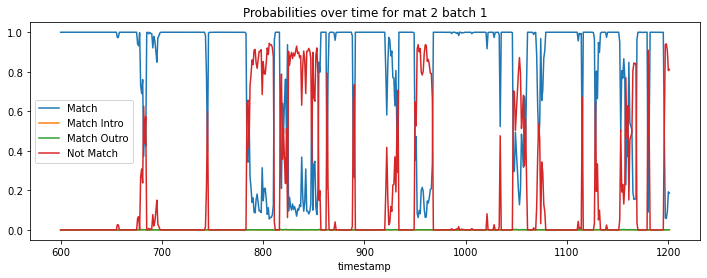}
    \caption{
        The results of applying fine-tuned YOLOv8 to two different mats sampled at the same ten-minute interval at 1 fps.
        Note that Mats 2 and 10 are raw camera footage due to technical issues with the stream.
    }
    \label{fig:match_classifier_time}
\end{figure}

In Figure \ref{fig:match_classifier_time}, we show the results of applying the classifier to two different mats at a ten-minute interval.
For videos containing the scoring overlay, we find a clear boundary between matches as given by the "Intro" and "Outro" classes.
However, there are several mats where the overlay is not present.
In this situation, the classifier struggles to differentiate between the "Match" and "Not Match" classes, causing fluctuations in the classification.

\subsection{Player and Referee Entity Detection}

We perform labeling of players and referees in three iterations, with the first iteration being the most time consuming.
After collecting approximately 700 unique annotations for the task, we fine-tune a detector and use the resulting model to generate bounding boxes for annotation.
The heuristic pre-annotator worked surprisingly well but struggled in occlusion situations.
In addition, class assignments based on color did not always work well, as the colors of the judo-gi are not always consistent.

When we fine-tuned the model, we found that the model better assigned the classes, but there were many duplicates.
However, situations with significant occlusions improved significantly.
In particular, ground-work bounding boxes were more accurate, covering the player's entire body behind the opponent.
After training the model with another 350 annotations, we find that the model can better filter out background people and objects.
In Table \ref{tab:lead_time_entity_detection}, we analyze the time it takes to generate an annotation.
Our first fine-tuned model, while accurate, was significantly slower than the heuristic-based method.
However, the semi-supervision process helped improve the results, leading to a faster annotation process.

\begin{table}[!t]
    \centering
    \caption{
        Model performance with semi-supervised label annotation.
        Each image typically has up to three distinct bounding boxes corresponding to a valid entity in the scene.
        An average lead time with 95\% confidence interval is computed with annotator statistics from Label Studio.
        There are significant differences introduced by successive versions of the fine-tuned model.
    }
    \label{tab:lead_time_entity_detection}
    \begin{tabular}{|l|r|r|r|r|}
    \hline
        \textbf{model} & 
        \textbf{images} & 
        \textbf{bboxes} & 
        \textbf{lead time (s)} & 
        \textbf{95\% CI} \\ \hline
    v0 & 0 & 0 & 9.66 & 0.73 \\ \hline
    v1 & 691 & 1,926 & 15.89 & 2.07 \\ \hline
    v2 & 357 & 868 & 10.13 & 1.26 \\ \hline
    v3 & 280 & 679 & N/A & N/A \\ \hline
    \end{tabular}
\end{table}





\subsection{Combat Phase Classification}

We calculate the distribution of each of the classes given the parent class in Table \ref{table:phase_label_distribution}.
We developed a pre-annotator for combat phase classification using computer vision features.
Initially, we used the aspect ratio of the detected player bounding boxes to distinguish between ground work and standing positions.
Table \ref{table:phase_distribution} highlights the challenge of accurately identifying standing versus ground work solely through aspect ratio thresholding of player bounding boxes.
The timer was then used to categorize phases into match and active classes (see Figure \ref{fig:example-timer-extraction}).
While pre-annotation aids in faster annotation, its performance falls short compared to other models, as shown in Table \ref{table:combat_phase_classification}.

\begin{table}[t]
    \centering
    \caption{
        Phase distribution of ground truth labels for combat phase classification.
    }
    \begin{tabular}{|c|c|c|c|}
        \hline
        P(match) & P(active) & P(standing) & count \\
        \hline
        1 & 1 & 1 & 106 \\
        1 & 1 & 0 & 177 \\
        1 & 0 & 0 & 155 \\
        0 & 0 & 0 & 132 \\
        \hline
    \end{tabular}
    \label{table:phase_label_distribution}
\end{table}

\begin{table}[!ht]
    \centering
    \caption{
        Distribution of match, active, and standing phases conditioned on dependent variables i.e. $P(\text{active} | \text{match}) = 0.646$.
    }
    \begin{tabular}{|c|r|r|r|}
        \hline
        & P(match) & P(active|match) & P(standing|active) \\ \hline
        \textbf{Annotated} & 0.768 & 0.646 & 0.242 \\ 
        \textbf{Heuristic} & 1.000 & 0.786 & 0.027 \\ \hline
    \end{tabular}
    \label{table:phase_distribution}
\end{table}

\begin{table}[!ht]
    \centering
    \caption{
        Top 5 combat phase classification model for each label.
    }
    \begin{tabular}{|c|c|c|c|}
        \hline
        \textbf{Label} & \textbf{Feature} & \textbf{Train F1} & \textbf{Test F1} \\
        \hline
        is match & fine tune v3 dctn & 0.96 & 0.66 \\
        is match & fine tune v2 dct d64 & 0.91 & 0.63 \\
        is match & fine tune v1 & 0.96 & 0.61 \\
        is match & fine tune v3 dct d16 & 0.88 & 0.61 \\
        is match & fine tune v2 & 0.95 & 0.58 \\
        is match & pre annotator & * & 0.77 \\
        \hline
        is active & fine tune v3 dct d8 & 0.61 & 0.78 \\
        is active & fine tune v2 dct d32 & 0.81 & 0.58 \\
        is active & fine tune v2 & 0.84 & 0.57 \\
        is active & pretrain yolov8n & 0.84 & 0.57 \\
        is active & fine tune v2 dctn & 0.84 & 0.53 \\
        is active & pre annotator & * & 0.45 \\
        \hline
        is standing & fine tune v3 dctn & 0.85 & 0.87 \\
        is standing & fine tune v2 dct d8 & 0.76 & 0.85 \\
        is standing & fine tune v3 dct d8 & 0.75 & 0.82 \\
        is standing & fine tune v3 dct d32 & 0.83 & 0.76 \\
        is standing & fine tune v2 dctn & 0.85 & 0.74 \\
        is standing & pre annotator & * & 0.37 \\
        \hline
    \end{tabular}
    \label{table:combat_phase_classification}
\end{table}

In Table \ref{table:combat_phase_classification}, successive versions of our fine-tuned detector embeddings perform better than the pretrained embeddings.
We often find that filtering the embeddings with the DCT improves the model's performance, both in F1 and training time.
In the DCT, we reduce (3, 12, 20) dimensions to k dimensions in the fine-tuned detector, and (80, 12, 20) in the pre-trained detector.
In the case of the "match" task, it effectively reduces the dimensional of the embeddings from 720 to 8 while improving the performance modestly, with a difference in 24 seconds using PySpark's logistic regression model, going from 32 to 8 seconds. 

\begin{table}[!ht]
    \centering
    \caption{
        Results of combat phase classification models using lagged features (DCT with 16 coefficients).
    }
    \begin{tabular}{|c|c|c|c|}
        \hline
        \textbf{Label} & \textbf{Feature} & \textbf{Train F1} & \textbf{Test F1} \\
        \hline
        is match & fine tune v3 lag1 & 0.90 & 0.62 \\
        is match & fine tune v3 & 0.88 & 0.61 \\
        is match & fine tune v2 & 0.92 & 0.56 \\
        is match & fine tune v3 lag2 & 0.92 & 0.56 \\
        is match & fine tune v2 lag3 & 0.94 & 0.53 \\
        \hline
        is active & fine tune v3 lag1 & 0.69 & 0.52 \\
        is active & fine tune v3 & 0.67 & 0.48 \\
        is active & fine tune v2 lag1 & 0.77 & 0.41 \\
        is active & fine tune v2 lag2 & 0.77 & 0.38 \\
        is active & fine tune v3 lag2 & 0.69 & 0.38 \\
        \hline
        is standing & fine tune v3 lag2 & 0.75 & 0.74 \\
        is standing & fine tune v3 lag3 & 0.79 & 0.74 \\
        is standing & fine tune v3 lag1 & 0.73 & 0.73 \\
        is standing & fine tune v3 & 0.79 & 0.72 \\
        is standing & fine tune v2 & 0.83 & 0.71 \\
        \hline
    \end{tabular}
    \label{table:combat_phase_classification_lag}
\end{table}

In Table \ref{table:combat_phase_classification_lag}, we see that the lagged features add some degree of performance to the model, but the differences are not necessarily significant.
Each additional lag reduces the number of labels available for testing, which means that our ranking of the models is not necessarily consistent.

\section{Discussion}

We explored three different annotation techniques for annotating judo matches from fixed-angle tournament footage.
We perform multi-class classification on full scenes, object detection for players and referees, and a multi-label classification for combat phases. 
Each task has a different labeling workflow and difficulty curve for annotators.
Labeling is by far the most time-consuming task of the pipeline, but it is also the most important because it is the only way to obtain ground truth data in this domain.

We start start with full-scene classification because it's conceptually simple and only requires a few seconds per scene to make a decision.
This annotation workflow is useful when there are regular patterns in the extracted frames, such as the software overlay on the footage.
There are situations that are ambiguous, so full-frame annotations do not make sense for scenes that require context.
For example, it is nearly indistinguishable to differentiate between players bowing in, players resetting to the center, or players bowing out.
For situations like this, having temporal context helps annotators label the situation.
In contrast, the video-interval labeling interface helps with situational ambiguity.
In our combat classification annotation setup, we sampled 30 second video clips from the dataset, which is enough to discriminate between various phases of combat.
The increased context comes with the trade-off of being more complex to label.

The object detection labeling workflow incorporated a semi-supervised labeling process, where we increased the amount of labeled data by using the predictions of a model to label more data.
In particular, we took advantage of the pre-annotation features of Label Studio to facilitate the labeling process.
We found this particularly useful for the entity detection model, which requires drawing three bounding boxes around the relevant entities to match in the foreground.
This process is time-consuming, taking about a dozen seconds per image. 

\section{Future Work}

We've explored annotation techniques for building models that can be incorporated into pipelines for automated analysis.
We can use the full-scene match classifier to segment video into individual matches, which would be useful when trying to find a specific match that occurred.
Additionally, if we build up a much larger ground-truth set of judo clips, we could build more complex models that incorporate temporal context in structured ways. 

We can also identify highlights of a match, potentially through the pose of the referee, which captures most of the information needed to understand the current state of the match.
A corresponding gesture accompanies every call made by the referee.
For example, the referee signals a winning score (ippon) by an arm straight up in the air while calling a match to a halt (matte), designated with an arm out with a palm facing the scorekeepers.
Referee poses could help with building an automated scoring system, where poses in matches with a referee are applied to techniques where the referee is not visible or in sparring situations.
Additionally, the pose carries signal about where scores are made, and thus could be used for automated highlight extraction.

Some other potential work includes building a system to identify the technique used to win a match.
Given proper labeling of match phases, we can truncate videos to all throwing techniques that lead to the termination of the match.
A compilation video can compress 100+ hours into a much shorter and more entertaining medium.
Additionally, the clips provide a valuable source of unlabeled training data that can serve to identify techniques by their official Kodokan classification.
Official Olympic scoring records note the technique used to win the match; smaller regional tournaments often do not because of the high effort on returns for bookkeeping. 
An automated statistical breakdown of techniques can be valuable for analyzing tournaments. 
It can also help label videos into functional groups as a learning resource for new judo players and referees.

\begin{table}[t]
\caption{
    A table of extracted statistics from recovered combat phase states.
    We choose statistics that validate the segmentation procedure and reproduce methodology from the literature.
}
\label{tab:tournament-statistics}
\begin{tabular}{|p{2.5cm}|p{5cm}|}
\hline
Statistic       & Description                             \\ \hline
Match Length    & The duration from timer start to reset. \\ \hline
Effort-Pause Ratio        & The ratio between active and non-active combat duration.    \\ \hline
Count of Throws & The total number of throws for points.  \\ \hline
Count of Pins and Submissions & The total number of pins or submissions for points.         \\ \hline
Standing-Ground Ratio     & The ratio of active combat spent standing vs on the ground. \\ \hline
\end{tabular}%
\end{table}

A complete model of combat phases would also let us compute and analyze statistics that are typically not recorded in tournaments due to the tediousness of data collection in the Table \ref{tab:tournament-statistics} by generating them from the combat phase states.
These statistics could be broken down by gender and weight class using the tournament software provided.

We also note some areas for potential multi-modality.
For our annotation purposes, we ignore the audio tracks associated with the video because many of the relevant cues relevant to analysis can be done visually.
However, the audio track is relevant because the referee will call out calls in addition to relevant hand gestures.
The score and timing setup also often includes a buzzer so that players know when the match is over.
Audio could play a role in the segmentation of the matches, especially if the signal is clear.
\section{Conclusion}

We present a set of annotation techniques to label fixed-angle videos of judo tournaments.
These are designed to segment sequential matches, and the extract combat phases from each match.
Semi-supervision is the primary technique required to facilitate modeling due to the lack of labeled data on judo competition combat phases.
We provide qualitative and quantitative results for our labeling process and our combat phase model, and discuss a future directions given an annotation system in place to gather ground-truth data.

\section*{Acknowledgments}
The authors would like to thank Cédric Pradalier and Stephanie Aravecchia for their support and guidance through this research.
We also acknowledge the use of the Georgia Tech Europe computer labs.

\bibliographystyle{plainnat}
\bibliography{main}

\end{document}